\documentclass[10pt,twocolumn,letterpaper]{article}

\usepackage[pagenumbers]{cvpr} % To force page numbers, e.g. for an arXiv version

\usepackage{multirow}
\usepackage{tabularx}
\newcolumntype{C}{>{\centering\arraybackslash}X}
\usepackage{booktabs}
\usepackage{amssymb}
\usepackage{pifont}

\definecolor{cvprblue}{rgb}{0.21,0.49,0.74}
\usepackage[pagebackref,breaklinks,colorlinks,allcolors=cvprblue]{hyperref}

\usepackage{xcolor}

\usepackage{amsthm}

\newtheorem{proposition}{Proposition}
\usepackage{multirow}

\def\paperID{528} % *** Enter the Paper ID here
\def\confName{3DV\xspace}
\def\confYear{2027\xspace}

\title{Guiding Image-to-3D Generation with Test-Time Partial Observations}

\author{Jerred Chen\\
University of Oxford\\
{\tt\small jerred.chen@cs.ox.ac.uk}
\and
Simon Weber\\
University of Oxford\\
{\tt\small simon.weber@cs.ox.ac.uk}
\and
Ronald Clark\\
University of Oxford\\
{\tt\small ronald.clark@cs.ox.ac.uk}
}

\begin{document}
\maketitle
\begin{abstract}

Image-to-3D models can generate visually compelling 3D assets from a single RGB image, but their geometry is often only loosely constrained by the available observations, limiting their use in applications that require geometric fidelity. In many real-world settings, however, partial geometric observations of the object may be available at test time. We introduce a training-free framework for incorporating such evidence into pretrained image-to-3D generative models without retraining or finetuning. To do this, we guide generation using a ray-consistent observation likelihood defined over the model's occupancy representation, combining surface occupancy and free-space evidence. Applied to SAM~3D and its multi-view extension, our approach substantially improves geometric fidelity across different levels of observability, as well as visual quality. Our results demonstrate that pretrained image-to-3D models can effectively integrate partial geometric observations through explicit test-time guidance, complementing their learned generative priors without modifying the underlying model.

\end{abstract}
    
\section{Introduction}\label{sec:intro}

\begin{figure}[t!]
\centering
\includegraphics[width=0.45\textwidth]{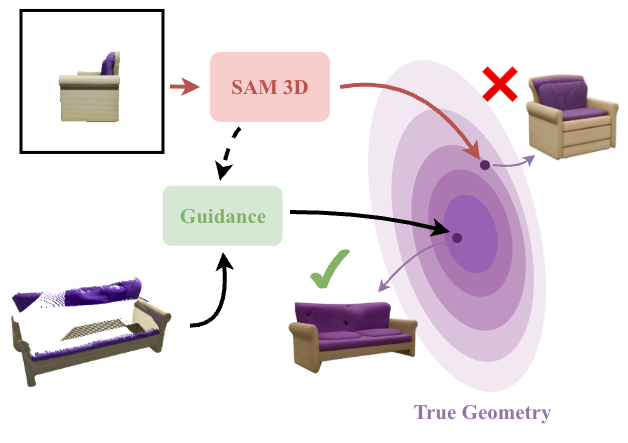}
\caption{Image-to-3D generation models like SAM 3D \cite{sam3dteam2025sam3d3dfyimages} can produce plausible outputs due to the model's learned prior, but available image observations often leave underlying geometry ambiguous: regions unseen from the input viewpoint (e.g. the sofa length) are reconstructed inconsistently with the true shape. With our proposed training-free guidance, the generation is steered towards the available information and recovers the correct geometry.}
\label{fig:failure-case}
\end{figure}

Image-to-3D generation models have recently achieved remarkable progress, producing high-quality 3D meshes and Gaussian splats from a single RGB image~\cite{sam3dteam2025sam3d3dfyimages, xiang2025structured}. Their impressive visual quality has made them attractive for applications ranging from digital content creation to robotics and mixed reality. However, this single-image setting also imposes a fundamental limitation: because a single image provides only ambiguous information about the underlying 3D shape, the generated geometry is largely determined by the model's learned prior rather than the true object geometry (see Figure \ref{fig:failure-case}). While the resulting assets are often visually plausible, they may exhibit incorrect depth, proportions, or hallucinated unseen surfaces.

This ambiguity is acceptable for creative applications, but it becomes problematic in settings where geometric accuracy is essential, such as robotics, digital twins, or augmented reality. In these scenarios, additional geometric information is often available at test time. Rather than relying solely on a single RGB image, one may have access to partial observations of the object, obtained for example from multiple viewpoints. These observations typically cover only a subset of the object, leaving the remaining geometry fundamentally ambiguous. The challenge is therefore to generate a complete 3D asset that simultaneously respects the observed geometry while leveraging the powerful shape prior learned by a pretrained image-to-3D model to plausibly complete the unobserved regions.

Existing approaches only partially address this problem. Multi-view extensions such as MV-SAM3D~\cite{li2026mv} improve consistency across several RGB images by modifying the generative model itself during sampling. However, they are tailored specifically to multiple RGB views and cannot naturally incorporate more general forms of geometric evidence such as partial point clouds or depth observations. More broadly, current image-to-3D models lack a generic mechanism for injecting arbitrary geometric constraints at inference time without retraining.
In this work, we address this limitation through a simple training-free guidance framework for image-to-3D generation. We guide a pretrained image-to-3D model using partial geometric observations available only at inference time, without modifying or finetuning the underlying model. Concretely, we interpret the model's canonical representation as an occupancy grid and construct an energy function directly from the observed geometry. This energy combines an occupancy term that encourages consistency with observed surfaces and a free-space term that prevents geometry from being generated where observations indicate empty space. Because the guidance acts solely on the sampling dynamics, it is entirely independent of the underlying image-to-3D architecture and can be readily integrated into existing models. In particular, it composes directly with multi-view extensions such as MV-SAM3D, combining the benefits of both approaches.

We evaluate our approach on SAM3D~\cite{sam3dteam2025sam3d3dfyimages} using partial geometric observations derived from multiple views. Our method substantially improves geometric fidelity while preserving the visual quality of the generated assets, demonstrating that pretrained image-to-3D foundation models can effectively combine learned generative priors with incomplete geometric evidence through purely test-time guidance.

In particular, our contributions are as follows:

\begin{itemize}
\item A training-free guidance framework for incorporating partial geometric observations into pretrained image-to-3D generation models at inference time

%\item A physically grounded, ray-consistent observation likelihood
\item A theoretical perspective for how test-time evidence deforms the instantaneous flow landscape

\item An occupancy-grid energy combining occupancy and free-space constraints built on a physically grounded, ray-consistent observation likelihood, allowing for enforcing geometric consistency and plausible completion of unseen regions

\item Extensive experiments demonstrating substantial improvements over state-of-the-art image-to-3D generation methods on partial observation reconstruction and novel view synthesis benchmarks.
\end{itemize}

\section{Related Works}
\label{sec:related_works}

% Surprisingly, very few works truly guide image-to-3d generation with external geometric observations, which is where our work contributes. Our review of related literature will focus on these two aspects: the guidance process in diffusion and flow matching on one hand, the image-to-3D generation process on the other hand.

% \subsection{Guidance in Diffusion and Flow Matching}
% The guidance in diffusion consists in controlling the model output 

% Historical foundations:
% \cite{song2020score}
% \cite{dhariwal2021diffusion}
% \cite{ho2022classifier}

% General training-free guidance:
% \cite{bansal2024universal}
% \cite{chung2022diffusion}
% \cite{ye2024tfg}
% \cite{yu2023freedom}

% Flow matching guidance:
% \cite{feng2025guidance}

While image-to-3D generation has progressed rapidly in recent years, surprisingly few works have investigated how to guide a pretrained image-to-3D model using external geometric observations available only at inference time. Existing methods either focus on improving the underlying generative model or on developing general guidance mechanisms for diffusion and flow matching. Our work lies at the intersection of these two directions: we leverage recent advances in training-free guidance to improve the geometric consistency of image-to-3D models without any retraining. 

\subsection{Guidance in Diffusion and Flow Matching} 
Guidance refers to mechanisms for steering the sampling process of a generative model towards samples satisfying a desired condition or constraint. In diffusion models, early forms of guidance relied either on an additional classifier trained to provide gradients towards a target class~\cite{dhariwal2021diffusion}, or on training the generative model itself to support both conditional and unconditional generation, as in classifier-free guidance~\cite{ho2022classifier}. These methods showed that the sampling trajectory can be modified at inference time to trade off sample fidelity and adherence to a desired condition. More recently, several works have generalized this idea beyond semantic conditioning.~\cite{bansal2024universal, chung2022diffusion, ye2024tfg, yu2023freedom} show that pretrained diffusion models can be guided by arbitrary differentiable objectives without retraining. Our work follows this line of research by designing a geometry-aware energy function tailored to image-to-3D generation. 
Finally, recent work has extended these ideas from diffusion models to flow matching. Feng \emph{et al.}~\cite{feng2025guidance} established the theoretical foundation of training-free guidance for flow-matching models, showing that guidance can be implemented through an additional velocity field during sampling. Building on this perspective, several methods have proposed practical guidance strategies for flow matching, including FlowChef~\cite{Patel2025FlowChefSO}, which steers the sampling trajectory using the gradient of a differentiable objective, and FlowDPS \cite{kim2025flowdps}, which extends diffusion posterior sampling to flow matching for inverse problems. Our work builds upon this growing line of research. Rather than proposing a new guidance algorithm, we design a geometry-aware energy tailored to image-to-3D generation from partial geometric observations.

%Finally, recent work has extended these ideas from diffusion models to flow matching. Feng \emph{et al.}~\cite{feng2025guidance} established the theoretical formulation of training-free guidance for flow matching models, showing that guidance can be implemented through an additional velocity field during sampling. We directly build upon this framework, but instead of proposing a new guidance algorithm, we introduce a geometry-guidance objective for image-to-3D generation based on sparse and noisy 3D observations. 

% \subsection{(Geometry-Constrained) Image-to-3D Generation}

% Diffusion-guided optimization:
% \cite{poole2022dreamfusion}
% \cite{lin2023magic3d}
% \cite{chen2023fantasia3d}
% \cite{wang2023prolificdreamer}

% Multi-view consistency:
% \cite{shi2024mvdream}
% \cite{li2024era3d}
% \cite{long2024wonder3d}
% \cite{li2026mv}

% Foundation models using geometry:
% \cite{xiang2025structured}
% \cite{boss2025sf3d}
% \cite{xiang2026native}

\subsection{Image-to-3D Generation} Recent advances in image-to-3D generation have largely been driven by diffusion and flow-based generative models capable of producing high-quality 3D assets from a single image. Early methods such as DreamFusion~\cite{poole2022dreamfusion}, Magic3D~\cite{lin2023magic3d}, Fantasia3D~\cite{chen2023fantasia3d}, and ProlificDreamer~\cite{wang2023prolificdreamer} formulate 3D generation as an optimization problem guided by pretrained diffusion models. These methods primarily rely on text or image guidance to synthesize plausible geometry, but do not incorporate external geometric observations during generation. Another line of work improves geometric quality by exploiting multiple input views. Methods such as MVDream~\cite{shi2024mvdream}, Era3D~\cite{li2024era3d}, Wonder3D~\cite{long2024wonder3d}, and the recent MV-SAM3D~\cite{li2026mv} enforce multi-view consistency during generation, leading to improved 3D reconstruction when several RGB observations are available. In contrast, our method is complementary to these approaches, as it leverages partial geometric observations rather than requiring additional RGB views. More recently, large-scale image-to-3D foundation models have demonstrated remarkable generation quality by learning structured latent representations of geometry and appearance~\cite{xiang2025structured,boss2025sf3d,xiang2026native}. In particular, SAM 3D~\cite{sam3dteam2025sam3d3dfyimages} can optionally condition its generation on geometric inputs such as point maps. However, we observe that these conditioning mechanisms do not always faithfully adhere to the provided geometry. In this work, we focus on SAM 3D and show that its geometric consistency can instead be substantially improved through training-free guidance at inference time, without retraining or modifying the underlying model.
\subsection{Partial 3D Conditioning}
Recent developments has shown promising directions in providing partial 3D information for image-to-3D generation. In particular, \cite{xia2026pointsto3d} propose training a model to inpaint the unobserved regions of a partial point cloud. \cite{hu2026axolotl3d} demonstrates a finetuned 3D generative model to take in additional condition tokens to complete the 3D generation. Contrary to these works, our method provides a straightforward recipe to guide the generation without any retraining. \cite{fedele2026spacecontrol} similarly shows a training-free approach by initializing the start of the generation with an interpolation between the partial 3D geometry encoding and Gaussian noise. We later show how applying our proposed guidance yields better results compared to the latent initialization.  

\section{Problem Formulation and Background}
Building on the motivation laid out in the introduction, we now place our objective on formal footing. We first distinguish \emph{learned conditioning}, which determines the pretrained generative prior, from \emph{test-time evidence}, which should constrain the generated geometry (\cref{subsec:problem}). We then review the flow-matching formulation of SAM~3D (\cref{subsec:flow_matching}), before showing in \cref{sec:method} how posterior guidance deforms the instantaneous landscape associated with its velocity field.

\subsection{Problem Formulation}\label{subsec:problem}
Let $c_I$ denote the standard conditioning information supplied to the pretrained image-to-3D model (e.g. image and object mask), and let $\mathcal{O}$ denote additional geometric evidence available only at test time, such as sparse point cloud. The pretrained model defines a conditional distribution $p_\theta(x\mid c_I)$ over plausible 3D assets $x$. Rather than requiring the learned conditioner to encode the observation faithfully, we treat $\mathcal{O}$ explicitly as evidence through an observation model $p(\mathcal{O}\mid x)$ that determines whether that asset agrees with the available measurements.

The desired posterior is therefore
\begin{align}
    q(x\mid c_I,\mathcal{O})
    &\propto
    p_\theta(x\mid c_I)\,p(\mathcal{O}\mid x)^{\beta} \\
    &=
    p_\theta(x\mid c_I)\exp\!\left[-\beta J_{\mathcal{O}}(x)\right],
    \label{eq:posterior}
\end{align}
where
\begin{equation}
    J_{\mathcal{O}}(x)=-\log p(\mathcal{O}\mid x)
    \label{eq:observation_energy}
\end{equation}
is the negative log-likelihood of the geometric observation and $\beta>0$ controls the strength of the evidence. This formulation cleanly separates the learned prior from the physical observation model.
%where $p_\theta(x\mid c_I)$ determines what constitutes a plausible 3D asset, while $p(\mathcal{O}\mid x)$ determines whether that asset agrees with the available measurements.

For the analysis below, we use $c$ to denote an arbitrary conditioner, e.g. either image-only conditioning $c_I$ or image-plus-geometry conditioning $c_{I,\mathcal{O}}$. Even when the same observation energy $J_{\mathcal{O}}$ is used, changing $c$ changes the pretrained flow and, consequently, the effective landscape on which the test-time guidance acts. 

Our method instantiates this framework for image-to-3D generation by introducing a guidance objective tailored to partial geometric observations. Before presenting our method and the new guided flow landscape in \cref{sec:method}, we briefly review how SAM 3D uses the flow matching for generating 3D samples.

\subsection{Flow Matching and SAM 3D Generation}\label{subsec:flow_matching}

\paragraph{Flow Matching.}
Starting from random noise, flow matching gradually transforms the noise into a realistic sample by following a learned velocity field, one small step at a time. More formally, let $p(\cdot,\cdot) : [0,1] \times \mathbb{R}^{d} \longrightarrow \mathbb{R}_{>0}$ be a probability path, i.e. for every $t \in [0,1]$, $p(t,\cdot)$ is a probability density over $\mathbb{R}^d$. We write $p_{0} := p(0,\cdot)$ for the initial (noise) distribution and $p_{1} := p(1,\cdot)$ for the target distribution.
Flow matching defines a vector field $v(\cdot,\cdot) : [0,1] \times \mathbb{R}^{d} \longrightarrow \mathbb{R}^{d}$ such that, starting from a sample $x_{0} \sim p_{0}$ and solving the ordinary differential equation
\begin{equation}\label{eq:ODE}
    \frac{d}{dt}x_{t} = v(t,x_t),
\end{equation}
the resulting trajectory satisfies $x_t \sim p(t,\cdot)$ for every $t \in [0,1]$; in particular, $x_1 \sim p_1$ is a clean sample from the target distribution. In this sense, $v$ drives the probability path $p$ from $p_0$ to $p_1$.

\paragraph{SAM 3D.}
SAM 3D instantiates this flow-matching formalism for 3D asset generation, and is decomposed, following \cite{xiang2025structured}, into two stages.
%Indeed,  SAM3D applies a flow-matching process to 3D asset generation. Following \cite{xiang2025structured}, SAM3D is decomposed in two stages. 
First, given image and mask, a geometric model predicts coarse shape $O$ and object pose $R,t,s$ (rotation,  translation, scale). Second, given image, mask and coarse shape, a texture model predicts the shape and texture $S,T$. The refined shape and texture latents are then decoded into Gaussian splats or meshes. Concretely, given a set of prediction modalities $\mathcal{M}$
%$\mathcal{M}=\{S,R,t,s\}$ 
and a set of conditioning modalities $c$
%$c=(I,M)$
, the model learns a conditional flow matching velocity field $v_{\theta}(t,x_t,c)$ \cite{lipman2022flow}:
\begin{align}
    \mathcal{L}_{\text{CFM}} = \sum_{m \in \mathcal{M}} \lambda_{m} \mathbb{E}_{\tau,x_{\tau}^{m}}\left[ \left\lVert v^{m} - v_{\theta}^{m}(\tau, x_{\tau}^{m},c) \right\rVert\right]
\end{align}
where the target velocity is the linear interpolation \cite{liu2022flow} 
\begin{align}
    v^{m} = x_{1}^{m} - x_{0}^{m} \,.
\end{align}
The goal is to generate $\{x_{1}^{m}\}_{m \in \mathcal{M}} \sim p(\mathcal{M} \vert c)$. In particular, during the first stage (geometry model), the prediction modalities are $\mathcal{M}=\{O,R,t,s\}$ and the conditioning modalities are $c=(I,M)$. During the second stage (texture and refinement model), $\mathcal{M}=\{S,T\}$ and $c=(O,I,M)$.
% At inference time, the generation proceeds by repeatedly applying the learned velocity: integrating \cref{eq:ODE} between $0$ and $1$ leads to the discrete update:
% \begin{align}\label{eq:update}
%     x_{t+\Delta t} = x_t + v_{\theta}(t,x_t,c) \cdot \Delta t \,.
% \end{align}
At inference time, the generation proceeds by repeatedly applying the learned velocity: integrating \cref{eq:ODE} between $0$ and $1$ leads to the discrete update
\begin{align}\label{eq:update}
    x_{t+\Delta t} = x_t + v_{\theta}(t,x_t,c) \cdot \Delta t \,,
\end{align}
which is applied iteratively from $t=0$ to $t=1$ to turn an initial noise sample $x_0$ into a final prediction $x_1$. We discuss how this update can be modified at inference time to bias generation towards samples satisfying a desired property.

\begin{figure*}[htbp!]
    \centering
    \includegraphics[width=0.9\textwidth]{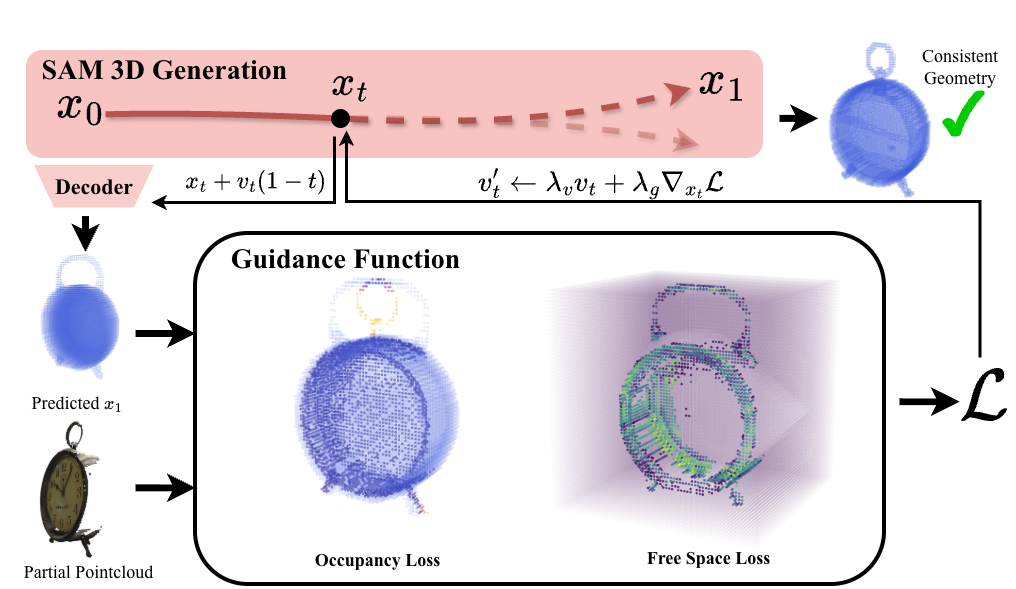}
    \caption{\textbf{Overview of Guided SAM~3D.} At each step $t$, the current state $x_t$ is decoded into a predicted clean sample $x_t + v_t(1-t)$, which is compared against the partial point cloud observation through an occupancy loss and a free-space loss, yielding a guidance signal $\mathcal{L}$. The velocity is corrected as $v'_t \leftarrow \lambda_v v_t + \lambda_g \nabla_{x_t}\mathcal{L}$ and fed back into the generation process, steering the trajectory towards a final sample $x_1$ with geometry consistent with the observations.}
    \label{fig:method}
\end{figure*}

\section{Method}\label{sec:method}

%The discrete update in \cref{eq:update} describes how SAM~3D generates a sample in the absence of any external information. Guidance \cite{bansal2024universal,feng2025guidance} lets us bias this step-by-step generation towards samples with a desired property -- in our case, agreement with the observations $\mathcal{O}$ -- without retraining the model, simply by adding an extra velocity term at every step. Let us first revisit the guidance process.
The discrete update in \cref{eq:update} describes how SAM~3D generates a sample from its learned conditional distribution. We extend this process to incorporate partial geometric evidence $\mathcal{O}$ at test time, without retraining the model. We first characterize how posterior guidance modifies the pretrained conditional flow, interpreting it as a deformation of an instantaneous flow landscape (\cref{subsec:landscape}). We then translate this perspective into a practical point-estimate guidance rule for the learned SAM~3D velocity (\cref{subsec:guidance}). Finally, we instantiate the observation energy through a ray-consistent likelihood, physically grounded, that explicitly accounts for observed surfaces and free space (\cref{subsec:ray_likelihood}).

\subsection{Guidance as a Deformation of the Flow Landscape}
\label{subsec:landscape}
To understand how test-time evidence modifies the pretrained generative process, we first characterize guidance \cite{bansal2024universal,feng2025guidance} at the level of the underlying flow. In particular, we show that, for the linear path used by SAM~3D, the conditional velocity admits an instantaneous potential whose deformation under posterior guidance can be made explicit. In particular, for the linear path
\begin{equation}
    x_t=(1-t)x_0+t x_1, \qquad x_0\sim\mathcal{N}(0,I),
    \label{eq:linear_path}
\end{equation}
the optimal conditional velocity can be written in terms of the score of its marginal $p_t^c(x)=p_t(x\mid c)$ as
\begin{equation}
    v_t^c(x)
    %v_{\theta}(t,x,c)
    = \frac{x}{t}
    + \frac{1-t}{t}\nabla_x\log p_t^c(x).
    \label{eq:native_velocity_score}
\end{equation}
Note that, at perfect-training limit, this velocity should converge to the velocity $v_\theta$ in \cref{eq:update}.
At each fixed $t>0$ the ideal velocity gives the instantaneous potential
\begin{align}
    J_v^c(t,x)
    &= -\frac{\lVert x\rVert^2}{2t}
       -\frac{1-t}{t}\log p_t^c(x) + K_{t},
    \qquad \\
    %v_{\theta}(t,x,c)&=-\nabla_x J_v^c(t,x).
    v_t^c(x)&=-\nabla_x J_v^c(t,x).
    \label{eq:native_potential}
\end{align}
We refer to $J_v^c$ as the \emph{native flow landscape}. Importantly, this landscape is defined instantaneously at each \(t\): sampling still follows the full time-dependent ODE rather than minimizing a fixed objective. However, this representation provides a convenient way to characterize how posterior guidance modifies the pretrained flow, as shown in the following proposition. 
% Define,
% \begin{equation}
%     h_t^c(x)
%     =
%     \mathbb{E}_{x_1\sim p(x_1\mid x_t=x,c)}
%     \left[\exp\!\left(-\beta J_{\mathcal{O}}(x_1)\right)\right].
%     \label{eq:ht}
% \end{equation}
% The corresponding intermediate marginal satisfies $q_t^c(x)\propto p_t^c(x)h_t^c(x)$. Its ideal velocity is therefore
% \begin{equation}
%     v_{vg}^c(t,x)
%     =
%     v_t^c(x)
%     +\frac{1-t}{t}\nabla_x\log h_t^c(x).
%     \label{eq:guided_velocity_exact}
% \end{equation}
% Equivalently, substituting $q_t^c$ for $p_t^c$ in \cref{eq:native_potential} gives the exact guided potential
% \begin{equation}
%     J_{vg}^c(t,x)
%     =
%     J_v^c(t,x)
%     -\frac{1-t}{t}\log h_t^c(x)
%     + C_t,
%     \label{eq:guided_potential_exact}
% \end{equation}
% where $C_t$ does not depend on $x$ and therefore has no effect on the velocity field. 
%We show in the following proposition that the posterior guidance can be viewed as an explicit deformation of the native flow landscape.

\begin{proposition}[Guided flow landscape]
    Let $J_v^{c}(t,x)$ denote the instantaneous potential associated with the conditional flow. Under posterior reweighting by geometric evidence $p(\mathcal{O}\mid x_1)^{\beta}$, the corresponding guided potential is:
    \begin{equation}
    J_{vg}^c(t,x)
    =
    J_v^c(t,x)
    -\frac{1-t}{t}\log h_t^c(x)
    + C_t \,,
    \label{eq:guided_potential_exact}
\end{equation}
where 
\begin{equation}
    h_t^c(x)
    =
    \mathbb{E}_{x_1\sim p(x_1\mid x_t=x,c)}
    \left[\exp\!\left(-\beta J_{\mathcal{O}}(x_1)\right)\right] \,.
    \label{eq:ht}
\end{equation}
Under the point-estimate approximation \cite{feng2025guidance,chung2022diffusion} $x_1 \approx x_t+(1-t)v_\theta(t,x_t,c)$, this becomes:
\begin{equation}
    J_{vg}^c(t,x_t)
    \approx
    J_v^c(t,x_t)
    + \eta_tJ_{\mathcal{O}}\!\left(x_{1}\right) + C_t\,,
    \label{eq:guided_potential_approx}
\end{equation}
with $\eta_t:=\beta\frac{1-t}{t}$.

\end{proposition}

Changing the conditioner changes both terms in \cref{eq:guided_potential_approx}: it changes the landscape $J_v^c$ through $p_t^c$, and it changes the observation landscape $\mathcal{E}_t^c(x_t) \equiv J_{\mathcal{O}}(x_1(x_t))$ through the condition-dependent map $x_1$. Also, note that $C_t$ is independent from $x$, and therefore has no effect on the velocity field. 
Proposition 1 thus reveals that test-time guidance acts as an explicit deformation of the pretrained flow landscape, rather than defining a separate generative process. 

\subsection{Guided SAM 3D}\label{subsec:guidance}
The landscape analysis above gives the exact posterior correction for the ideal flow. For the learned SAM 3D velocity, we use the corresponding point-estimate approximation at every sampling step, similarly as prior training-free guidance methods \cite{feng2025guidance,chung2022diffusion}. Specifically, we predict the clean endpoint
\begin{align}\label{eq:clean}
    \tilde{x}_1(x_t)
    =
    x_t+v_\theta(t,x_t,c)(1-t),
\end{align}
and evaluate the observation energy
\begin{equation}
    \mathcal{E}_t^c(x_t;\mathcal{O})
    =
    J_{\mathcal{O}}\!\left(\tilde{x}_1(x_t)\right).
    \label{eq:practical_energy}
\end{equation}
The guidance velocity is then
\begin{equation}
    g(t,x_t,c)
    =
    -\lambda_g(t)\nabla_{x_t}\mathcal{E}_t^c(x_t;\mathcal{O}),
    \label{eq:guidance_velocity}
\end{equation}
where $\lambda_g(t)$ contains constant arising from the exact posterior-guidance expression as well as the guidance-strength schedule used in practice. The guided velocity is
\begin{equation}
    v^g(t,x_t,c)
    =
    v_\theta(t,x_t,c)+g(t,x_t,c),
    \label{eq:guided_velocity}
\end{equation}
and the discrete sampling update becomes
\begin{align}\label{eq:sample_new_distribution}
    x_{t+\Delta t}
    =
    x_t+
    \left(v_\theta(t,x_t,c)
    -\lambda_g(t)\nabla_{x_t}\mathcal{E}_t^c(x_t;\mathcal{O})\right)\Delta t.
\end{align}
%This is the same point-estimate energy-guidance mechanism used by prior training-free guidance methods \cite{feng2025guidance,chung2022diffusion} but 
Let us now describe our novel contribution, that is to instantiate $J_{\mathcal{O}}$ as a physically interpretable likelihood of partial geometric observations and to show how the resulting guidance interacts with the pretrained conditional flow landscape.

\subsection{Ray-Consistent Observation Likelihood}
\label{subsec:ray_likelihood}
We now instantiate the observation energy $J_{\mathcal{O}}$ from the given measurement geometry. We derive this energy from the likelihood of an observed camera ray under the occupancy field obtained from the predicted clean sample.

Let $\pi_v(x)\in[0,1]$ denote the predicted probability that voxel $v$ is occupied. Consider an observed ray $r$ whose measured surface lies in voxel $s_r$, and let $\mathcal{F}_r$ denote the set of voxels traversed by the ray before reaching $s_r$. The observation implies that every voxel in $\mathcal{F}_r$ is empty and that $s_r$ is occupied. Under a Bernoulli occupancy model, the likelihood of the ray is
\begin{equation}
    p(\mathcal{O}_r\mid x)
    =
    \pi_{s_r}(x)
    \prod_{v\in\mathcal{F}_r}\left(1-\pi_v(x)\right).
    \label{eq:ray_likelihood}
\end{equation}
Assuming conditional independence across observed rays gives
\begin{equation}
    p_{\mathrm{ray}}(\mathcal{O}\mid x)
    =
    \prod_{r\in\mathcal{R}}p(\mathcal{O}_r\mid x),
    \label{eq:all_ray_likelihood}
\end{equation}
and therefore the negative log-likelihood
\begin{equation}
    J_{\mathrm{ray}}(x;\mathcal{O})
    =
    -\sum_{r\in\mathcal{R}}\log\pi_{s_r}(x)
    -\sum_{r\in\mathcal{R}}\sum_{v\in\mathcal{F}_r}
    \log\left(1-\pi_v(x)\right).
    \label{eq:ray_nll}
\end{equation}
The two terms have a direct physical interpretation as the first is a surface-hit likelihood, while the second is aray-survival/free-space likelihood. A depth observation is probable only when the ray remains empty until the measured surface and becomes occupied at the measured surface.

\begin{figure*}[ht!]
    \centering
    \includegraphics[width=0.9\textwidth]{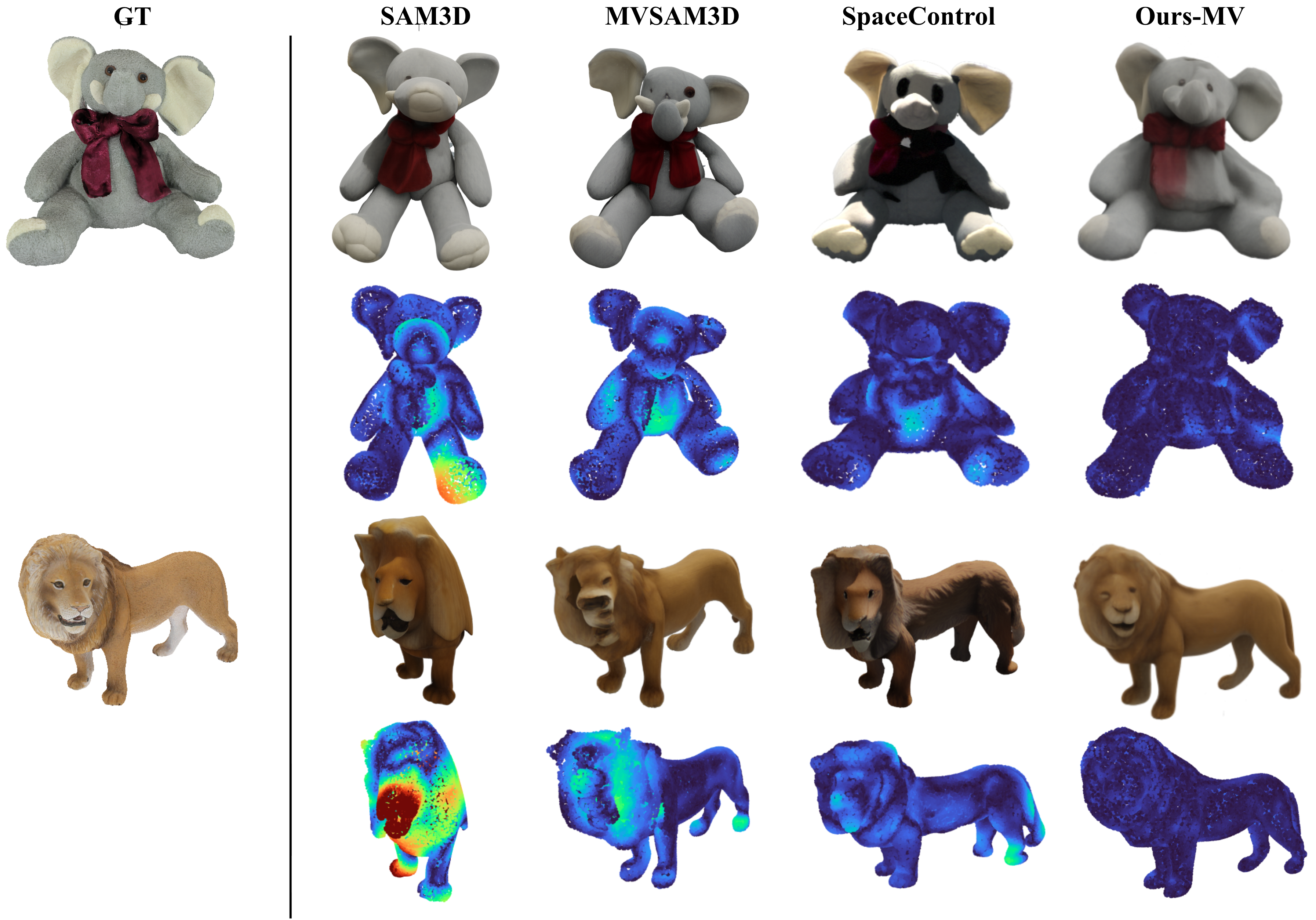}
    \caption{\textbf{Qualitative comparison on two examples under high observability.} Each example consists of two rows: the generated Gaussian splat and predicted-to-ground-truth one-sided distance heatmap (blue indicates low error, red indicates high error). Our method leads to substantially more accurate geometry while preserving the visual appearance of the generated asset.}
    \label{fig:high}
\end{figure*}
\begin{table*}[ht!]
\centering
\begin{tabularx}{\textwidth}{@{} l | CCC | CCC | CCC @{}}
\toprule
\multirow{2}{*}{Method}
& \multicolumn{3}{c|}{\textbf{Low Observability}}
& \multicolumn{3}{c|}{\textbf{Medium Observability}}
& \multicolumn{3}{c}{\textbf{High Observability}} \\
\cmidrule(lr){2-4}
\cmidrule(lr){5-7}
\cmidrule(lr){8-10}
& CD $\downarrow$ & SD $\downarrow$ & LPIPS $\downarrow$
& CD $\downarrow$ & SD $\downarrow$ & LPIPS $\downarrow$
& CD $\downarrow$ & SD $\downarrow$ & LPIPS $\downarrow$ \\
\midrule

SAM 3D \cite{sam3dteam2025sam3d3dfyimages}
& 16.8 & 53.5 & 0.326
& -- & -- & --       % 2 views
& -- & -- & -- \\    % 5 views

+ matching 3D cond.
& 15.0 & 50.4 & 0.326
& 15.0 & 50.4 & 0.326       % 2 views
& 15.0 & 50.4 & 0.326 \\    % 5 views

MV-SAM3D \cite{li2026mv}
& 17.5 & 54.9 & 0.326
& 10.1 & 41.4 & 0.300  % 2 views
& 9.60 & 39.6 & 0.282 \\ % 5 views

+ matching 3D cond.
& 15.0 & 50.4 & 0.326
& 10.2 & 41.5 & 0.290   % 2 views
& 8.85 & 37.3 & 0.286 \\ % 5 views

% SpCtrl., $\tau_0=3$ \cite{fedele2026spacecontrol}
% & 29.4 & 110. & 0.381  % 1 view
% & 21.7 & 92.1 & 0.367  % 2 views
% & 17.6 & 77.8 & 0.337 \\ % 5 views

SpaceControl \cite{fedele2026spacecontrol}
& 10.3 & 55.9 & 0.294  % 1 view
& \textbf{2.43} & \underline{22.5} & 0.265  % 2 views
& \textbf{0.70} & \textbf{12.3} & 0.241 \\ % 5 views

\midrule

Ours
& \textbf{7.84} & \textbf{36.5} & \textbf{0.286}
& 3.43 & 24.1 & \underline{0.245}                 % 2 views
& 1.45 & 16.4 & \underline{0.220} \\ % 5 views

Ours + MV
& \underline{8.11} & \underline{37.2} & \underline{0.289}
& \underline{2.83} & \textbf{22.0} & \textbf{0.224}                 % 2 views
& \underline{1.02} & \underline{15.7} & \textbf{0.201} \\   % 5 views

\bottomrule
\end{tabularx}

\caption{Quantitative comparison of 3D reconstruction accuracy using
Chamfer Distance (CD) and Ground-Truth Sided Distance (SD) for the geometry, and LPIPS for the rendering.
Our method are competitive or outperforms the corresponding baselines across
the reported evaluation metrics and numbers of input views. Note that
all distance metrics are scaled by $10^{3}$.}
\label{tab:metrics_comparison}
\end{table*}

\begin{figure*}[ht!]
    \centering
    \includegraphics[width=0.95\textwidth]{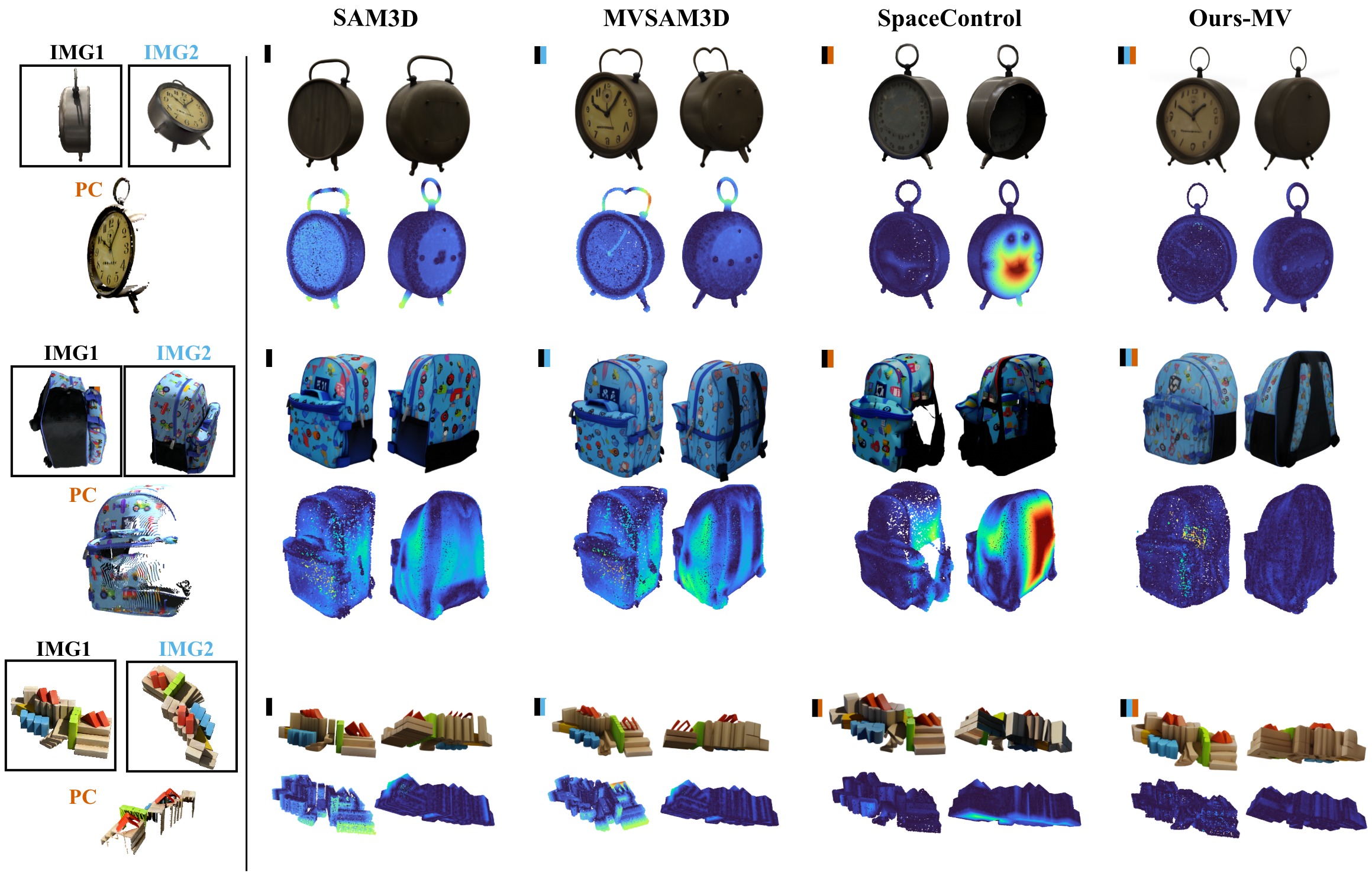}
    \caption{\textbf{Qualitative comparison under medium observability.} The left column shows the available RGB images (IMG1, IMG2) and the guidance partial point cloud (PC). For each method, we show the generated Gaussian splat, the predicted-to-ground-truth sided distance error heatmap (left), and ground-truth-to-predicted heatmap (right). Incorporating partial geometric observations through our training-free guidance consistently produces more accurate geometry than the baselines, while remaining complementary to multi-view generation. The colored tags indicate the inputs available to each method: black for IMG1, blue for IMG2, orange for partial point cloud (PC).}
    \label{fig:medium}
\end{figure*}

\section{Experiments}

\paragraph{Baselines.} We compare our method against the following:

\begin{itemize}
    \item \textit{SAM 3D} \cite{sam3dteam2025sam3d3dfyimages} is the state-of-the-art image-to-3D generation model, especially in occluded environments. We evaluate using the default proposed hyperparameters.
    \item \textit{MV-SAM3D} \cite{li2026mv} proposes a multi-view extension on top the original SAM 3D method by performing a weighted average across all of the velocity predictions at each timestep during the generation. We use the default hyperparameters and proposed entropy-based weighting.
    \item \textit{SpaceControl \cite{fedele2026spacecontrol}} is a recent method that inputs geometric primitives into encoded latents and initializes the sampling process with these latents. To keep the input modalities consistent, we opt to not use the text conditioning in the provided SpaceControl demo and use TRELLIS' original image-to-3D backbone. We initialize at generation step $\tau_0 =6$, the default in its evaluations.
\end{itemize}
    
\paragraph{Benchmark.} We evaluate on the object-centric dataset GSO-30 \cite{downs2022_gsodataset, kong2024eschernet}. GSO-30 is a suite of 30 real-world objects with a full 360$^\circ$ render obtained from the original GSO dataset. We evaluate in three settings: high, medium, and low observability. Each observability setting is determined by randomly selecting 5, 2, and 1 image(s) respectively, such that the images in each lower observability setting is strictly a subset of the higher observability's images. In each setting, we simulate RGB-D measurements by projecting the ground-truth mesh onto the selected views and only using the points visible from those views. In the case of a method accepting multiple RGB images such as MV-SAM3D, all of the selected view images are used as input. Both SAM 3D and MV-SAM3D use MoGe \cite{wang2025moge} to estimate pointmaps that are embedded and used for conditioning. To keep the measurements consistent, we also evaluate against passing in the same RGB-D measurements into the conditioning embedder, which we denote as ``matching 3D cond." in Table \ref{tab:metrics_comparison}. SAM 3D only receives one input image, so we only report for the low observability case, which is . We additionally evaluate on extending our method to MV-SAM3D, in which we perform the velocity weighting scheme proposed and additionally apply classifier guidance.

\paragraph{Metrics.} We evaluate on the following geometry and novel view metrics:
\begin{itemize}
    \item Chamfer Distance measures the bidirectional distance between the source pointcloud and the nearest neighbor in the target pointcloud. We report the squared Chamfer Distance, scaled by $10^{3}$.
    \item Ground-truth Sided Distance measures specifically the distance between each ground-truth point to its nearest neighbor in the prediction pointcloud. We report this part of the Chamfer Distance calculation to emphasize any inaccuracies the prediction may have with gaping holes or missing regions. We again report this scaled by $10^3$.
    \item LPIPS \cite{zhang2018perceptual} computes the differences between each layer activation of VGG-16, which corresponds well with human perceptual features. We evaluate the mean LPIPS across 8 randomly sampled novel views that are at least 30 degrees away from the selected source views. Each baseline method receives the same selected novel views.
\end{itemize}
For the geometric metrics, we follow the SAM 3D evaluation protocol by independently normalizing both the ground truth and predicted meshes between [-1, 1] and performing ICP to align with the ground truth before computing the metrics. We randomly sample 30,000 points on each mesh. To evaluate LPIPS, we use the decoded Gaussian splat from each method to render the novel views.

\paragraph{Results.} \label{subsec:results} Table \ref{tab:metrics_comparison} shows our quantitative results. In all observabilities, we greatly surpass the performance of both SAM 3D and MV-SAM3D. In low observability, we outperform all baselines in geometric and novel view metrics, and we maintain competitive geometric performance compared to SpaceControl in medium and high observabilities while consistently outperforming in novel view synthesis. 

Figures \ref{fig:high} and \ref{fig:medium} visualize the Gaussian splat predictions from each method and the corresponding error heatmap compared to the ground truth. These figures further exemplify the importance of explicit guidance of partial observations compared to only using RGB images or the latent initialization approach as in SpaceControl. Since SAM 3D does not take in any additional inputs, the resulting shape is ambiguous and is unable to correct the imperfect geometry from any additional signals. Using more RGB images can help as seen in MV-SAM3D, but is not sufficient and still results in inaccuracies in the predicted geometry. The latent initialization done in SpaceControl performs well under high observability, but the ground-truth-to-predicted heat maps in Figure \ref{fig:medium} shows a major limitation with more limited views -- without enough coverage, SpaceControl is prone to missing large portions of the geometry, as seen in the missing backs of the alarm clock and backpack, and the empty bottom surface of the wooden blocks. In contrast, our method consistently adheres to the geometry while maintaining more faithful visual appearances in novel views.
\section{Ablations}

\begin{table}[t]
    \centering
    \caption{Ablations on the guidance design decisions. Run across the GSO30 dataset with high observability, single-view setup.}
    \label{tab:ablation}
    \begin{tabular}{lcc}
        \toprule
        \textbf{Ablated Feature}
        & \textbf{CD} $\downarrow$ & \textbf{LPIPS} $\downarrow$ \\
        \midrule
        Baseline ($\lambda_g = 1,\lambda_f = 1$, steps=50)                               & 15.1 & 0.341 \\
        (+) $\lambda_g = 4$             & 4.01 & 0.287 \\
        (+) $\lambda_f = 2$         & 3.54 & 0.275 \\
        (+) Increase steps to 200                 & \textbf{0.98} & 0.228 \\
        % (+) Replace w/ conditional guidance       & 1.28 & 0.222 \\
        % (-) Remove pointmap conditioning          & 1.32 & 0.223 \\
        (+) Guidance weight cooldown              & 1.45 & \textbf{0.220} \\
        % + Early exit                            & 1.76 & 0.223 \\
        \bottomrule
    \end{tabular}
\end{table}

Table \ref{tab:ablation} shows what decision decisions were helpful for successfully applying classifier guidance to SAM 3D. The ablation is performed across the entire GOS30 dataset with high observability without any multiview images for simplicity. We found that without any additional weighting to the guidance gradient, the method performed on par with vanilla SAM 3D, even after normalizing the gradient to match the magnitude of the original velocity. Only by significantly upweighting the guidance gradient did we find major improvements into adhering to the geometry as well as more realistic renderings in novel views. We also found that the performance could be further improved by upweighting the free-space loss, as the magnitude of the occupancy loss largely dominates the total loss. Increasing the number of sampling steps yielded major improvement in the performance as well. We hypothesize that the generation is not accustomed to the application of the test-time guidance, which requires more steps to accommodate. While we obtained good geometric accuracy from this point, we observed that the visual quality of the resulting 3D objects were unsatisfactory, with unsmooth surfaces along with some artifacts such as holes or floaters. Applying a guidance weight cooldown schedule reduced its geometric performance, but this improved the appearances as well as the novel view renderings. 
\section{Conclusion}

%We presented a simple training-free framework for incorporating partial geometric observations into pretrained image-to-3D generation models at inference time. By interpreting the model's canonical representation as an occupancy grid, we introduced a geometry-aware energy combining occupancy and free-space constraints that guides the sampling process towards the observed geometry while allowing the pretrained generative prior to plausibly complete the unseen regions. Applied to SAM~3D, our approach consistently improves geometric fidelity while preserving visual quality, demonstrating that powerful image-to-3D foundation models can effectively integrate external geometric evidence without any retraining or finetuning.
%Our work also has several limitations. First, we assume that the partial observations are accurate and expressed in the model's canonical coordinate system. Handling noisy observations or unknown coordinate frames would require more robust guidance objectives or joint alignment strategies, which we leave for future work. Second, our guidance relies on occupancy-grid representations and therefore inherits their finite spatial resolution. Finally, while we demonstrate the approach on SAM~3D and its multi-view extension, extending the same idea to other image-to-3D generation models remains an interesting direction for future research.

We presented a training-free framework for incorporating partial geometric observations into pretrained image-to-3D generation models at inference time. Beyond the practical guidance mechanism, we interpreted posterior guidance as a deformation of the pretrained conditional flow landscape. Building on this perspective, we derived a ray-consistent observation likelihood in the model's occupancy representation, combining surface occupancy and free-space evidence to steer generation towards the observed geometry while retaining the learned prior to complete unobserved regions. Applied to SAM~3D and its multi-view extension, our approach consistently improves geometric fidelity across different levels of observability while preserving visual quality.
Our experiments further highlight that explicitly enforcing geometric evidence through guidance can be substantially more effective than providing the same information solely through the model's learned conditioning pathway. Together, these results show that partial geometric evidence can be incorporated explicitly at test time to complement, rather than replace, the powerful priors learned by image-to-3D foundation models.
Our work also has several limitations. First, we assume that the partial observations are accurate and expressed in the model's canonical coordinate system. Handling noisy observations or unknown coordinate frames would require more robust observation models or joint alignment strategies. Second, our geometric guidance operates on an occupancy-grid representation and therefore inherits its finite spatial resolution. More generally, our formulation opens the possibility of designing observation likelihoods for other forms of geometric evidence and intermediate 3D representations. While we demonstrate our approach on SAM~3D and its multi-view extension, extending explicit geometric guidance to other image-to-3D generative models remains another promising direction.
{
    \small
    \bibliographystyle{ieeenat_fullname}
    \bibliography{main}
}

\clearpage
\pagenumbering{roman}
\appendix

\setcounter{page}{1}
\maketitlesupplementary

\noindent This supplemental material is organized as follows:\\
\noindent \textbf{\cref{sec:proof_proposition_1}} presents a proof for Proposition 1. \\
\noindent \textbf{\cref{sec:extended}} shows an extended set of visualizations.

\section{Proof of Proposition 1}\label{sec:proof_proposition_1}

Let us define
\begin{equation}
    h_t^c(x)
    =
    \mathbb{E}_{x_1\sim p(x_1\mid x_t=x,c)}
    \left[\exp\!\left(-\beta J_{\mathcal{O}}(x_1)\right)\right].
    \label{eq:ht}
\end{equation}
The corresponding intermediate marginal satisfies $q_t^c(x)\propto p_t^c(x)h_t^c(x)$. Its ideal velocity is therefore
\begin{equation}
    v_{vg}^c(t,x)
    =
    v_t^c(x)
    +\frac{1-t}{t}\nabla_x\log h_t^c(x).
    \label{eq:guided_velocity_exact}
\end{equation}
Equivalently, substituting $q_t^c$ for $p_t^c$ in \cref{eq:native_potential} gives the exact guided potential
\begin{equation}
    J_{vg}^c(t,x)
    =
    J_v^c(t,x)
    -\frac{1-t}{t}\log h_t^c(x)
    + C_t,
    \label{eq:guided_potential_exact}
\end{equation}
where $C_t$ does not depend on $x$ and therefore has no effect on the velocity field. 

\section{Extended Visualizations}\label{sec:extended}

\begin{figure*}[ht!]
    \centering
    \includegraphics[width=0.95\textwidth]{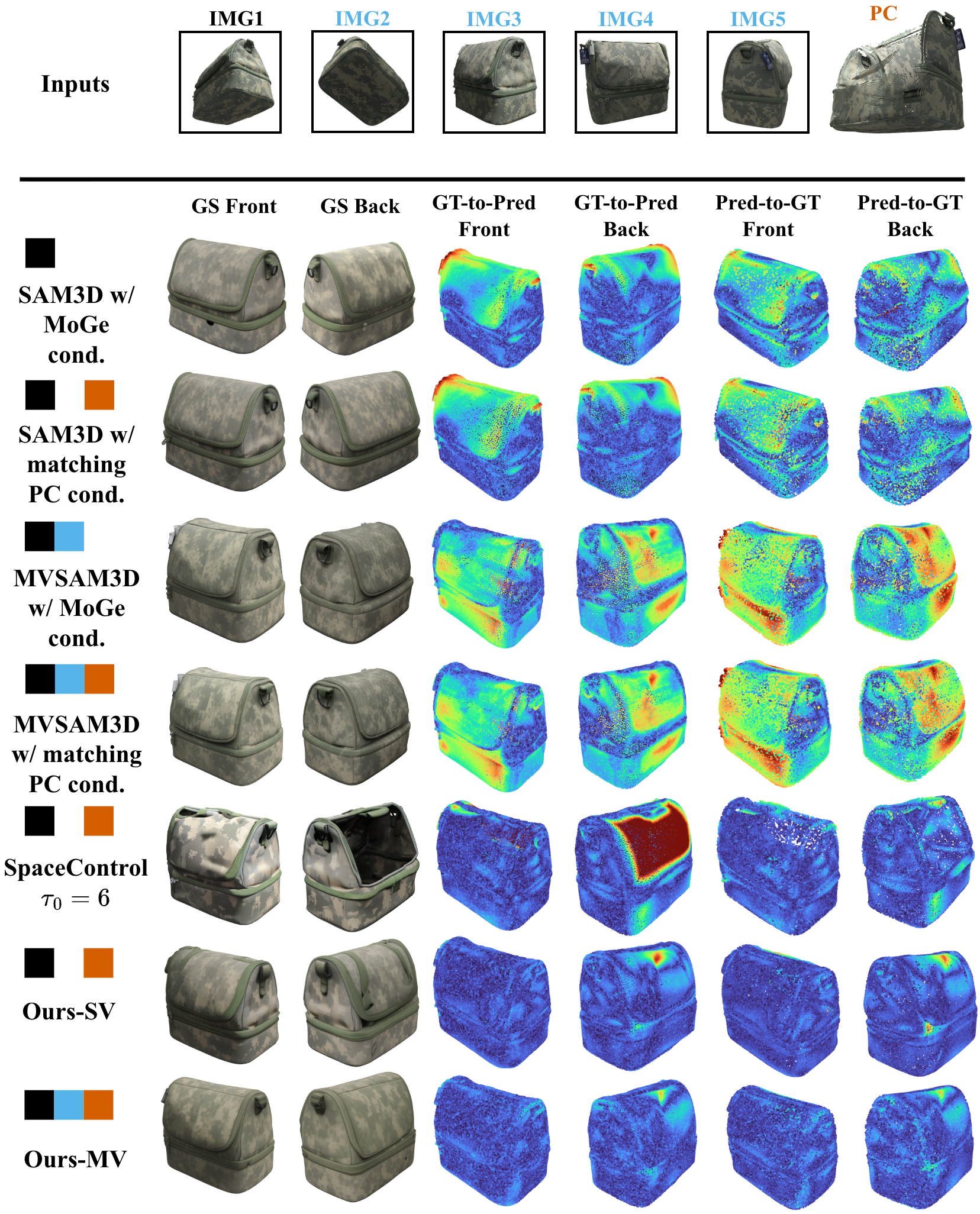}
    \caption{\textbf{Qualitative comparison of the ``lunch bag" object under high observability.} At the top, we show all potential image inputs and the partial point cloud. Each baseline shows the fronts and backs of the Gaussian splat (GS), the GT-to-prediction sided error heatmap, and the prediction-to-GT sided error heatmap. Each method is color-coded with the corresponding accepted inputs.}
    \label{fig:extended1}
\end{figure*}

\begin{figure*}[ht!]
    \centering
    \includegraphics[width=0.95\textwidth]{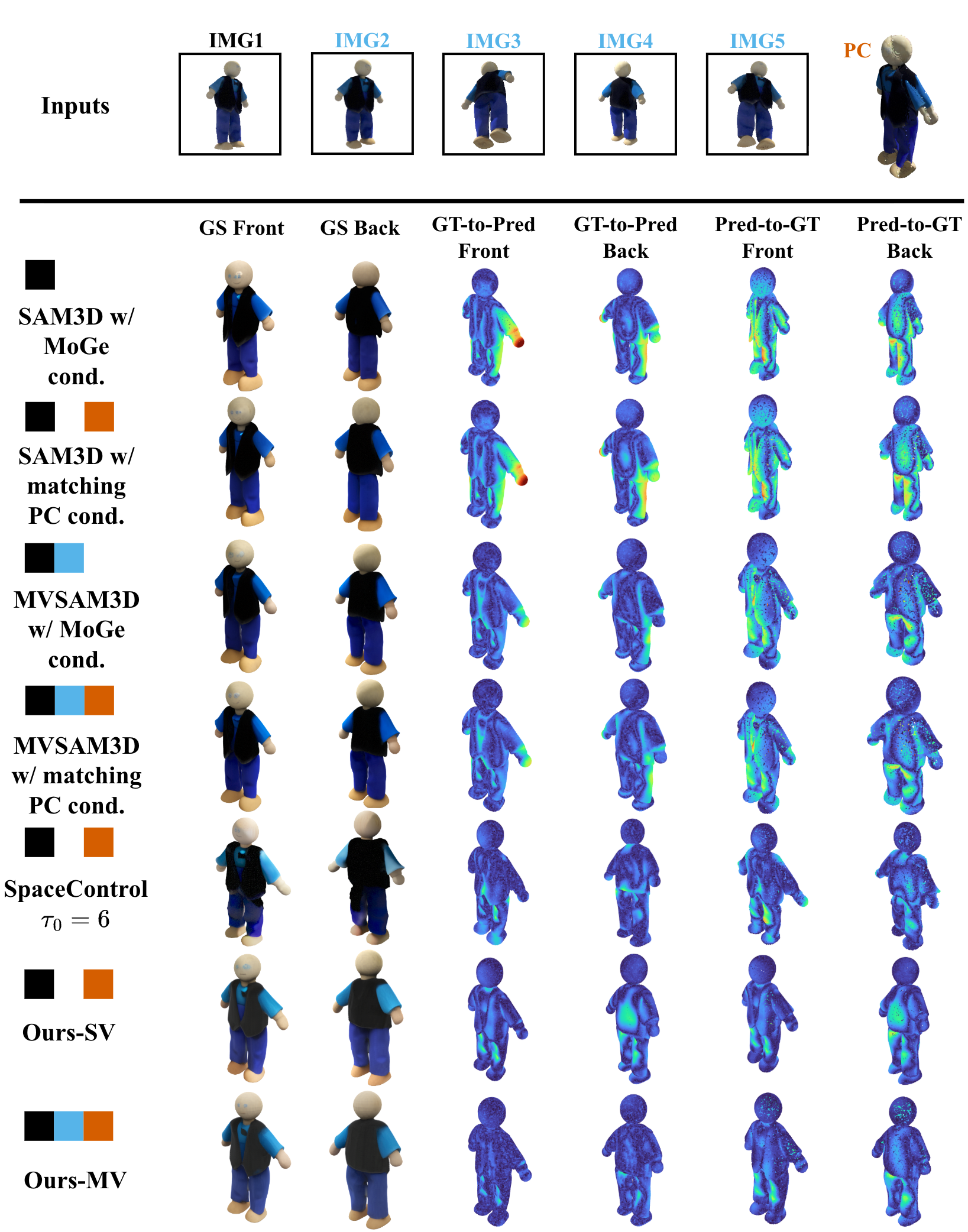}
    \caption{\textbf{Qualitative comparison of the ``grandfather" object under high observability.} At the top, we show all potential image inputs and the partial point cloud. Each baseline shows the fronts and backs of the Gaussian splat (GS), the GT-to-prediction sided error heatmap, and the prediction-to-GT sided error heatmap. Each method is color-coded with the corresponding accepted inputs.}
    \label{fig:extended2}
\end{figure*}

Figures \ref{fig:extended1} and \ref{fig:extended2} show more detailed comparisons between all baselines. Notably, we include the visualizations for SAM 3D and MV-SAM 3D using its default MoGe conditioning as well as inputting the partial point cloud for pointmap conditioning, labelled ``matching PC cond.". Even when using the same partial point cloud as conditioning, we have found only minor improvements in the performance, which is also observed quantitatively in Table \ref{tab:metrics_comparison}. We additionally show the visualizations from our method Ours-SV (single-view). Figure \ref{fig:extended1} in particular demonstrates the effectiveness of applying our method to single-view SAM 3D, outperforming all other baselines. SpaceControl here is unable to generate the unobserved regions, highlighting the importance of explicitly applying guidance over the latent initialization. Ours-MV further improves the performance using additional multi-view information to constrain the generation.

% \section{Rationale}
% \label{sec:rationale}
% % 
% Having the supplementary compiled together with the main paper means that:
% % 
% \begin{itemize}
% \item The supplementary can back-reference sections of the main paper, for example, we can refer to \cref{sec:intro};
% \item The main paper can forward reference sub-sections within the supplementary explicitly (e.g. referring to a particular experiment); 
% \item When submitted to arXiv, the supplementary will already included at the end of the paper.
% \end{itemize}
% % 
% To split the supplementary pages from the main paper, you can use \href{https://support.apple.com/en-ca/guide/preview/prvw11793/mac#:~:text=Delete%20a%20page%20from%20a,or%20choose%20Edit%20%3E%20Delete).}{Preview (on macOS)}, \href{https://www.adobe.com/acrobat/how-to/delete-pages-from-pdf.html#:~:text=Choose%20%E2%80%9CTools%E2%80%9D%20%3E%20%E2%80%9COrganize,or%20pages%20from%20the%20file.}{Adobe Acrobat} (on all OSs), as well as \href{https://superuser.com/questions/517986/is-it-possible-to-delete-some-pages-of-a-pdf-document}{command line tools}.

\end{document}